\documentclass[11pt,a4paper]{article}

\usepackage[hyperref]{emnlp2020}
\usepackage{times}
\usepackage{latexsym}

\usepackage[utf8]{inputenc} % was utf8x before; not sure if it matters

\usepackage[T2A,LAE,T1]{fontenc}
\usepackage[russian,arabic,USenglish]{babel}
\usepackage{CJKutf8}

\usepackage{graphicx}
\usepackage[figuresleft]{rotating}
\usepackage{amsmath}
\usepackage{amssymb}
\usepackage{microtype}
\usepackage{hyperref}
\usepackage{cleveref}

\usepackage{multirow}
\usepackage{tabulary}
\usepackage{array}

\usepackage{caption}
\usepackage{subcaption}
\usepackage{booktabs}
\usepackage[multiple]{footmisc}
\usepackage{amsmath}

\aclfinalcopy % Uncomment this line for the final submission
\def\aclpaperid{3216} %  Enter the acl Paper ID here

\newcommand{\entity}[1]{\texttt{#1}}
\newcommand{\ientity}[2]{\entity{#1}\textsuperscript{\href{https://www.wikidata.org/wiki/#2}{#2}}}

\newcommand*{\TR}{TR2016\textsuperscript{hard}}

\newcolumntype{L}[1]{>{\raggedright\let\newline\\\arraybackslash\hspace{0pt}}m{#1}}

\title{Entity Linking in 100 Languages}

\author{Jan A. Botha \\ Google Research \\
  \texttt{jabot@google.com} \\\And
  Zifei Shan \\ Google Research \\
  \texttt{zifeishan@gmail.com} \\\And
  Daniel Gillick \\ Google Research \\
  \texttt{dgillick@google.com}}

\date{}

\begin{document}
\maketitle
\begin{abstract}
We propose a new formulation for \emph{multilingual entity linking}, where language-specific mentions resolve to a language-agnostic Knowledge Base.
We train a dual encoder in this new setting, building on prior work with improved feature representation, negative mining, and an auxiliary entity-pairing task, to obtain a single entity retrieval model that covers 100+ languages and 20~million entities.
The model outperforms state-of-the-art results from a far more limited cross-lingual linking task.
Rare entities and low-resource languages pose challenges at this large-scale, so we advocate for an increased focus on zero- and few-shot evaluation.
To this end, we provide \textbf{Mewsli-9}, a large new multilingual dataset\footnote{\url{http://goo.gle/mewsli-dataset}} matched to our setting, and show how frequency-based analysis provided key insights for our model and training enhancements.
\end{abstract}

\section{Introduction}
Entity linking (EL) fulfils a key role in grounded language understanding:
Given an ungrounded entity mention in text, the task is to identify the entity’s corresponding entry in a Knowledge Base (KB).
In particular, EL provides grounding for applications like Question Answering \citep{fevry2020entities} (also via Semantic Parsing \citep{shaw2019generating}) and Text Generation \citep{puduppully2019data}; it is also an essential component in knowledge base population \citep{shen2014entity}. Entities have played a growing role in representation learning. For example, entity mention masking led to greatly improved fact retention in large language models \citep{guu2020realm,roberts2020much}.

But to date, the primary formulation of EL outside of the standard monolingual setting has been \emph{cross-lingual}: link mentions expressed in one language to a KB expressed in another \citep{mcnamee-etal-2011-cross,tsai-roth-2016-cross,sil2018neural}. The accompanying motivation is that KBs may only ever exist in some well-resourced languages, but that text in many different languages need to be linked. 
Recent work in this direction features progress on low-resource languages \cite{zhou_tacl2020},
zero-shot transfer \cite{sil-florian-2016-one,rijhwani2019zero,zhou2019towards}
and scaling to many languages \cite{pan-etal-2017-cross}, but commonly assumes
a single primary KB language and a limited KB, typically English Wikipedia.

We contend that this popular formulation limits the scope of EL in ways that are artificial and inequitable.

First, it artificially simplifies the task by restricting the set of viable entities and reducing the variety of mention ambiguities. Limiting the focus to entities that have English Wikipedia pages understates the real-world diversity of entities.
Even within the Wikipedia ecosystem, many entities only have pages in languages other than English.
These are often associated with locales that are already underrepresented on the global stage.
By ignoring these entities and their mentions, most current modeling and evaluation work tend to side-step under-appreciated challenges faced in practical industrial applications, which often involve KBs much larger than English Wikipedia, with a much more significant zero- or few-shot inference problem.

Second, it entrenches an English bias in EL research that is out of step with the encouraging shift toward \emph{inherently multilingual} approaches in natural language processing, enabled by advances in representation learning \cite{johnson2017google,pires-etal-2019-multilingual,conneau-etal-2020-unsupervised}.

Third, much recent EL work has focused on models that rerank entity candidates retrieved by an alias table \citep{fevry2020empirical}, an approach that works well for English entities with many linked mentions, but less so for the long tail of entities and languages.

To overcome these shortcomings, this work makes the following key contributions:

\begin{itemize}
    \item Reformulate entity linking as inherently multilingual: link mentions in 104 languages to entities in WikiData, a language-agnostic KB.
    \item Advance prior dual encoder retrieval work with improved mention and entity encoder architecture and improved negative mining targeting.
    \item Establish new state-of-the-art performance relative to prior cross-lingual linking systems, with one model capable of linking 104 languages against 20 million WikiData entities.
    \item Introduce \textbf{Mewsli-9}, a large dataset with nearly 300,000 mentions across 9 diverse languages with links to WikiData. The dataset features many entities that lack English Wikipedia pages and which are thus inaccessible to many prior cross-lingual systems.
    \item Present frequency-bucketed evaluation that highlights zero- and few-shot challenges with clear headroom, implicitly including low-resource languages without enumerating results over a hundred languages. 
\end{itemize}

\section{Task Definition}
\emph{Multilingual Entity Linking} (MEL) is the task of linking an entity mention $m$ in
some context language $l^{\textrm{c}}$ to the corresponding entity $e\in V$ in a \emph{language-agnostic KB}. That is, while the KB may include textual information (names, descriptions, etc.) about each entity in one or more languages, we make no prior assumption about the relationship between these KB languages $L^{\textrm{kb}}=\{l_1, \dots, l_k\}$ and the mention-side language: $l^{\textrm{c}}$ may or may not be in $L^{\textrm{kb}}$.

This is a generalization of \emph{cross-lingual EL} (XEL), which is concerned with the case 
where $L^{\textrm{kb}}=\{l'\}$ and $l^{\textrm{c}} \neq l'$.
Commonly, $l'$ is English, and $V$ is moreover limited to the set of entities that express features in $l'$.

\subsection{MEL with WikiData and Wikipedia}
As a concrete realization of the proposed task, we use WikiData \cite{vrandevcic2014wikidata} as our KB:
it covers a large set of diverse entities,
is broadly accessible and actively maintained,
and it provides access to entity features in many languages.
WikiData itself contains names and short descriptions,
but through its close integration with all Wikipedia editions,
it also connects entities to rich descriptions (and other features) drawn from the corresponding language-specific Wikipedia pages.

Basing entity representations on features of their Wikipedia pages has been a common approach 
in EL \cite[e.g.][]{sil-florian-2016-one,francis-landau-etal-2016-capturing,gillick-etal-2019-learning,wu2019zeroshot}, but we will need to generalize this to include multiple Wikipedia pages with possibly redundant features in many languages.

\subsubsection{WikiData Entity Example}\label{sec:description_example}
Consider the WikiData Entity \ientity{Sí Ràdio}{Q3511500}, a now defunct Valencian radio station. Its KB entry references Wikipedia pages in three languages, which contain the following descriptions:\footnote{We refer to the first sentence of a Wikipedia page as a description because it follows a standardized format.}

\begin{itemize}
 \item (Catalan) \emph{\textbf{Sí Ràdio} fou una emissora de ràdio musical, la segona de Radio Autonomía Valenciana, S.A. pertanyent al grup Radiotelevisió Valenciana.}
 \item (Spanish) \emph{\textbf{Nou Si Ràdio} (anteriormente conocido como Sí Ràdio) fue una cadena de radio de la Comunidad Valenciana y emisora hermana de Nou Ràdio perteneciente al grupo RTVV.}
 \item (French) \emph{\textbf{Sí Ràdio} est une station de radio publique espagnole appartenant au groupe Ràdio Televisió Valenciana, entreprise de radio-télévision dépendant de la Generalitat valencienne.}
% \item[] \emph{gloss: Sí Ràdio is a Spanish public radio station belonging to the Ràdio Televisió Valenciana group, a radio and television company dependent on the Valencian Generalitat.}
\end{itemize}

Note that these Wikipedia descriptions are not direct translations, and contain some name variations.
We emphasize that this particular entity would have been completely out of scope in the standard cross-lingual task \citep{tsai-roth-2016-cross}, because it does not have an English Wikipedia page.

In our analysis, there are millions of WikiData entities with this property, meaning the standard setting skips over the substantial challenges of modeling these (often rarer) entities, and disambiguating them in different language contexts.
Our formulation seeks to address this.

\subsection{Knowledge Base Scope}
Our modeling focus is on using \emph{unstructured textual} information for entity linking,
leaving other modalities or structured information as areas for future work.
Accordingly, we narrow our KB to the subset of entities that have descriptive text available:
We define our entity vocabulary $V$ as all WikiData items that have an associated Wikipedia page in \emph{at least one language}, independent of the languages we actually model.\footnote{More details in Appendix~\ref{appsec:dataprep}.}
This gives 19,666,787 entities,
\emph{substantially more than in any other task settings we have found}: the KB accompanying the entrenched TAC-KBP 2010 benchmark \citep{ji2010overview} has less than a million entities, and although English Wikipedia continues to grow, recent work using it as a KB still only contend with roughly 6~million entities \citep{fevry2020empirical,zhou_tacl2020}.
Further, by employing a simple rule to determine the set of viable entities, we avoid potential selection bias based on our desired test sets or the language coverage of a specific pretrained model.

\subsection{Supervision}
We extract a supervision signal for MEL by exploiting the hyperlinks that editors place on Wikipedia pages, taking the anchor text as a linked mention of the target entity.
This follows a long line of work in exploiting hyperlinks for EL supervision \cite{bunescu-pasca-2006-using,singh12:wiki-links,logan-etal-2019-baracks},
which we extend here by applying the idea to extract a large-scale dataset of 684 million mentions in 104 languages, linked to WikiData entities.
This is at least six times larger than datasets used in prior English-only linking work \citep{gillick-etal-2019-learning}.
Such large-scale supervision is beneficial for probing the quality attainable with current-day high-capacity neural models.

\section{Mewsli-9 Dataset}
We facilitate evaluation on the proposed multilingual EL task by releasing a matching dataset that covers a diverse set of languages and entities.

\textbf{Mewsli-9} (\emph{\textbf{M}ultilingual Entities in N\textbf{ews}, \textbf{li}nked)} contains 289,087 entity mentions appearing in 58,717 originally written news articles from WikiNews,
linked to WikiData.%
\footnote{\url{www.wikinews.org}, using the 2019-01-01 snapshot from \url{archive.org}}

The corpus includes documents in nine languages, representing five language families and six orthographies.%\footnote{Arabic, German, English, Spanish, Persian, Japanese, Serbian, Tamil, Turkish}
\footnote{Mewsli-9 languages \textcolor{gray}{(code, family, script)}:
Japanese \textcolor{gray}{(`ja', Japonic, ideograms)};
German   \textcolor{gray}{(`de', Indo-European (IE), Latin)};
Spanish  \textcolor{gray}{(`es', IE, Latin)};
Arabic   \textcolor{gray}{(`ar', Afro-Asiatic,} \textcolor{gray}{Arabic)};
Serbian  \textcolor{gray}{(`sr', IE, Latin \& Cyrillic)};
Turkish  \textcolor{gray}{(`tr', Turkic, Latin)};
Persian  \textcolor{gray}{(`fa', IE, Perso-Arabic)};
Tamil    \textcolor{gray}{(`ta', Dravidian, Brahmic)};
English  \textcolor{gray}{(`en', IE, Latin)}.}
Per-language statistics appear in \autoref{tab:wn_corpus}.
Crucially, 11\% of the 82,162 distinct target entities in \mbox{Mewsli-9}
\emph{do not have English Wikipedia pages},
thereby setting a restrictive upper bound on performance attainable by a standard XEL system focused on English Wikipedia entities.\footnote{As of 2019-10-03.}
Even some English documents may contain such mentions, such as the Romanian reality TV show, \ientity{Noră pentru mama}{Q12736895}.

WikiNews articles constitute a somewhat different text genre from our Wikipedia training data: The articles do not begin with a formulaic entity description, for example, and anchor link conventions are likely different. We treat the full dataset as a test set, avoiding any fine-tuning or hyperparameter tuning, thus allowing us to evaluate our model’s robustness to domain drift.

\mbox{Mewsli-9} is a drastically expanded version of the English-only WikiNews-2018 dataset by \newcite{gillick-etal-2019-learning}.
Our automatic extraction technique trades annotation quality for scale and diversity, in contrast to the MEANTIME corpus based on WikiNews \cite{minard-etal-2016-meantime}.
\mbox{Mewsli-9} intentionally stretches the KB definition beyond English Wikipedia, unlike VoxEL \cite{rosales2018voxel}.
Both MEANTIME and VoxEL are limited to a handful of European languages.

\begin{table}
\small
\centering
\begin{tabular}{rrrrr} \toprule
& & & \multicolumn{2}{c}{\textbf{Entities}} \\ \cmidrule(lr){4-5}
\textbf{Lang.} & \textbf{Docs} & \textbf{Mentions} & Distinct & $\notin$ EnWiki \\ \midrule
ja & 3,410 & 34,463 & 13,663 & 3,384 \\
de & 13,703 & 65,592 & 23,086 & 3,054 \\
es & 10,284 & 56,716 & 22,077 & 1,805 \\
ar & 1,468 & 7,367 & 2,232 & 141 \\
sr & 15,011 & 35,669 & 4,332 & 269 \\
tr & 997 & 5,811 & 2,630 & 157 \\
fa & 165 & 535 & 385 & 12 \\
ta & 1,000 & 2,692 & 1,041 & 20 \\
en & 12,679 & 80,242 & 38,697 & 14 \\ \cmidrule(lr){2-5} 
   & 58,717 & 289,087 & 82,162 & 8,807 \\ \midrule 
en$'$  & 1801 & 2,263 & 1799 & 0  \\ \bottomrule
\end{tabular}
\caption{Corpus statistics for Mewsli-9, an evaluation set we introduce for multilingual entity linking against WikiData.
%The final column shows how many entities do not have a dedicated English Wikipedia page.
Line en$'$ shows statistics for English WikiNews-2018, by \newcite{gillick-etal-2019-learning}.
\label{tab:wn_corpus}}
\end{table}

\begin{figure*}
  \centering
  \includegraphics[width=\textwidth]{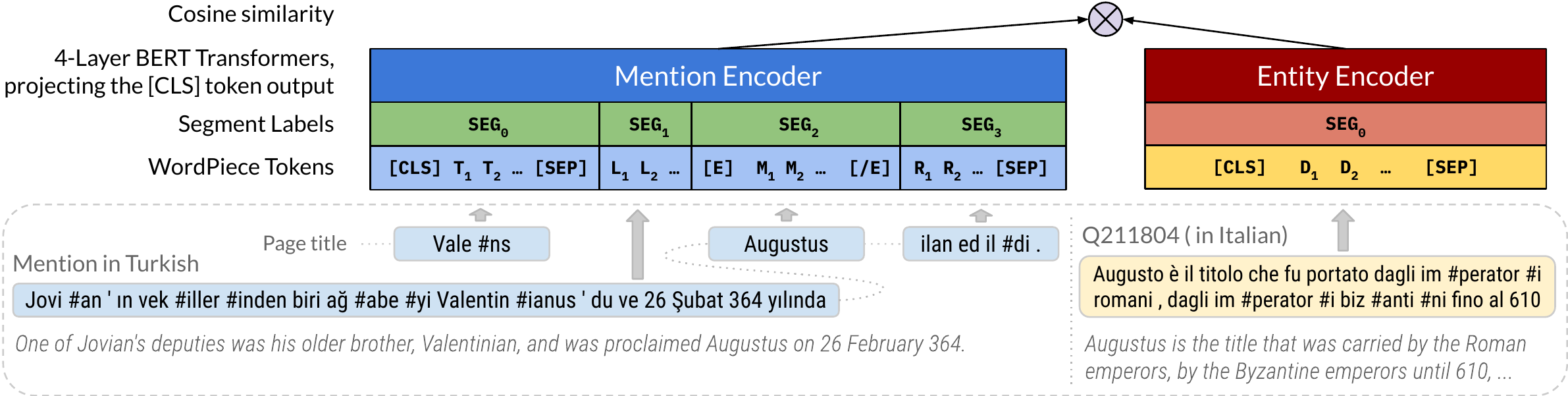}
  \caption{
  \small Dual Encoder \textbf{Model F} diagram. The input to the \textit{Mention Encoder} is a sequence of WordPiece tokens that includes the document title ($T_i$), context immediately left of the mention ($L_i$), the mention span ($M_i$) demarcated by [E] and [/E] markers, and context immediately right of the mention ($R_i$). Segment labels ($SEG_i$) are also used to distinguish the input segments. The input to the (Model F) \textit{Entity Encoder} is simply the WordPiece tokens in the entity description ($D_i$). As usual, embeddings passed to the first transformer layer are the sum of positional embeddings (not pictured here), the segment embeddings, and the WordPiece embeddings. The example shows a Turkish mention of \ientity{Augustus}{Q211804} paired with its Italian description. \label{fig:architecture}} 
\end{figure*}

\section{Model}
Prior work showed that a dual encoder architecture can encode entities and contextual mentions in a dense vector space to facilitate efficient entity retrieval via nearest-neighbors search \citep{gillick-etal-2019-learning,wu2019zeroshot}.
We take the same approach.
The dual encoder maps a mention-entity pair $(m,e)$ to a score:
\begin{equation}\label{eq:cosine}
    s(m,e) = \frac{\phi(m)^T \psi(e)}{\|\phi(m)\| \|\psi(e)\|},
\end{equation}
where $\phi$ and $\psi$ are learned neural network encoders that encode their arguments as $d$-dimensional vectors ($d$=300, matching prior work).

Our encoders are BERT-based Transformer networks \cite{vaswani2017attention,devlin-etal-2019-bert}, which we initialize from a pretrained multilingual BERT checkpoint.%
\footnote{\url{github.com/google-research/bert} \texttt{multi\_cased\_L-12\_H-768\_A-12}}
For efficiency, we only use the first 4 layers, which results in a negligible drop in performance relative to the full 12-layer stack. 
The WordPiece vocabulary contains 119,547 symbols covering the top 104 Wikipedia languages by frequency---this is the language set we use in our experiments.

\subsection{Mention Encoder}
The mention encoder $\phi$ uses an  input representation that is a combination of \emph{local context} (mention span with surrounding words, ignoring sentence boundaries) and simple \emph{global context} (document title). The document title, context, and mention span are marked with special separator tokens as well as identifying token type labels (see \autoref{fig:architecture} for details). Both the mention span markers and document title have been employed in related work \citep{agarwal2020entity,fevry2020empirical}.  We use a maximum sequence length of 64 tokens similar to prior work \citep{fevry2020empirical}, up to a quarter of which are used for the document title. The CLS token encoding from the final layer is projected to the encoding dimension to form the final mention encoding.

\subsection{Entity Encoders}
We experiment with two entity encoder architectures. The first, called \textbf{Model F}, is a featurized entity encoder that uses a fixed-length text description (64 tokens) to represent each entity (see \autoref{fig:architecture}). The same 4-layer Transformer architecture is used---without parameter sharing between mention and entity encoders---and again the CLS token vector is projected down to the encoding dimension. Variants of this entity architecture were employed by \newcite{wu2019zeroshot} and \newcite{logeswaran-etal-2019-zero}.

The second architecture, called \textbf{Model E} is simply a QID-based embedding lookup as in \newcite{fevry2020empirical}.
This latter model is intended as a baseline. \emph{A priori}, we expect Model E to work well for common entities, less well for rarer entities, and not at all for zero-shot retrieval. We expect Model F to provide more parameter-efficient storage of entity information and possibly improve on zero- and few-shot retrieval.

\subsubsection{Entity Description Choice}\label{sec:description_heuristic}
There are many conceivable ways to make use of entity descriptions from multiple languages.
We limit the scope to using one primary description per entity, thus obtaining a single coherent text fragment to feed into the Model F encoder.

We use a simple data-driven selection heuristic that is based on observed entity usage:
Given an entity $e$,
let $n_e(l)$ denote the number of mentions of $e$ in documents of language $l$,
and $n(l)$ the global number of mentions in language $l$ across all entities.
From a given source of descriptions---first Wikipedia and then WikiData---we order the candidate descriptions $(t_e^{l_1},t_e^{l_2},\dots )$ for  $e$ first by the per-entity distribution $n_e(l)$ and then by the global distribution $n(l)$.%
\footnote{The candidate descriptions (but not $V$) are limited to the 104 languages covered by our model vocabulary---in general, both Wikipedia and WikiData cover more than 300 languages.} For the example entity in Section~\ref{sec:description_example}, this heuristic selects the Catalan description because $9/16$ training examples link to the Catalan Wikipedia page.%, while the other seven link to the Spanish page.

\subsection{Training Process}
In all our experiments, we use an 8k batch size with in-batch sampled softmax \citep{gillick2018end}.
Models are trained with Tensorflow \citep{abadi2016tensorflow} using the Adam optimizer \cite{kingma2015adam,Loshchilov2019DecoupledWD}.
All BERT-based encoders are initialized from a pretrained checkpoint, but the Model E embeddings are initialized randomly.
%These training settings were mostly adapted from prior work. 
We doubled the batch size until no further held-out set gains were evident and chose the number of training steps to keep the training time of each phase under one day on a TPU. Further training would likely yield small improvements.
See Appendix~\ref{appsec:training} for more detail.

\section{Experiments}
We conduct a series of experiments to gain insight into the behavior of the dual encoder retrieval models under the proposed MEL setting, asking:
\begin{itemize}
\itemsep-0.2em
\item What are the relative merits of the two types of entity representations used in Model E and Model F (embeddings vs. encodings of textual descriptions)?% (\autoref{sec:eval_design_choices})
\item Can we adapt the training task and hard-negative mining to improve results across the entity frequency distribution?
\item Can a single model achieve reasonable performance on over 100 languages while retrieving from a 20 million entity candidate set?% (\autoref{sec:linking_in_100}
\end{itemize}

\subsection{Evaluation Data}
We follow \newcite{upadhyay-etal-2018-joint} and evaluate on the ``hard'' subset of the Wikipedia-derived test set introduced by
\newcite{tsai-roth-2016-cross} for cross-lingual EL against English Wikipedia, \textbf{\TR}.
This subset comprises mentions for which the correct entity did not appear as the top-ranked item in their alias table, thus stress-testing a model's ability to generalize beyond mention surface forms. 

Unifying this dataset with our task formulation and data version requires mapping its gold entities from the provided, older Wikipedia titles to newer WikiData entity identifiers (and following intermediate Wikipedia redirection links).
This succeeded for all but 233/42,073 queries in \TR{}---our model receives no credit on the missing ones.

To be compatible with the pre-existing train/test split, we excluded from our training set all mentions appearing on Wikipedia pages in the full TR2016 test set.
This was done for all 104 languages, to avoid cross-lingual overlap between train and test sets.
This aggressive scheme holds out 33,460,824 instances, leaving our final training set with 650,975,498 mention-entity pairs. \autoref{fig:heldout_104} provides a break-down by language.

\subsection{Evaluating Design Choices}\label{sec:eval_design_choices}

\subsubsection{Setup and Metrics}
In this first phase of experiments we evaluate design choices by reporting the \emph{differences} in Recall@100 between two models at a time, for conciseness.
Note that for final system comparisons, it is standard to use Accuracy of the top retrieved entity (R@1), but to evaluate a dual encoder retrieval model, we prefer R@100 as this is better matched to its likely use case as a candidate generator.

Here we use the \TR{} dataset, as well a portion of the 104-language set held out from our training data, sampled to have 1,000 test mentions per language.
(We reserve the new \mbox{Mewsli-9} dataset for testing the final model in \autoref{sec:eval_wikinews9}.)

Reporting results for 104 languages is a challenge. 
To break down evaluation results by entity
frequency bins, we partition a test set according to the frequency of its gold entities as observed in the training set.
This is in line with recent recommendations for finer-grained evaluation in EL \citep{waitelonis-gerbil2016,ilievski-etal-2018-systematic}.

We calculate 
metrics within each bin, and report macro-average over bins. 
This is a stricter form of the label-based macro-averaging sometimes used,
but better highlights the zero-shot and few-shot
cases. We also report micro-average metrics, computed over the entire dataset, without binning.

%%%%%%%%%%%%%%%%%%%%%%%%%%%%%%%%%%%%%%%%%%%%%%%%%%%
%%%% Three combined.
%%%%%%%%%%%%%%%%%%%%%%%%%%%%%%%%%%%%%%%%%%%%%%%%%%%%
\begin{table*}
\centering
\begin{tabular}{l rr rr rr} \toprule
 & \multicolumn{2}{c}{\bf (a)}  & \multicolumn{2}{c}{\bf (b)}  & \multicolumn{2}{c}{\bf (c)}   \\
 \cmidrule(lr){2-3}
 \cmidrule(lr){4-5}
 \cmidrule(lr){6-7}
\textbf{Bin} & holdout & \TR{} & holdout & \TR{} & holdout & \TR{} \\ 
\midrule
$[0, 1)$ & +0.842 & +0.380 & +0.009 & +0.093 & +0.044 & +0.144 \\
$[1, 10)$ & +0.857 & +0.814 & +0.018 & +0.037 & +0.051 & +0.031 \\
$[10, 100)$ & +0.211 & +0.191 & +0.012 & +0.024 & +0.006 & -0.019 \\
$[100, 1k)$ & -0.010 & -0.031 & +0.007 & +0.019 & -0.005 & -0.015 \\
$[1k, 10k)$ & -0.018 & -0.051 & +0.008 & +0.011 & -0.003 & -0.007 \\
$[10k, +)$ & -0.009 & -0.089 & +0.004 & +0.003 & -0.002 & -0.013 \\ \midrule
micro-avg & +0.018 & +0.008 & +0.006 & +0.017 & -0.001 & -0.006 \\
macro-avg & +0.312 & +0.202 & +0.010 & +0.031 & +0.015 & +0.020 \\ \bottomrule
\end{tabular}
\caption{R@100 differences between pairs of models:
(a) model F (featurized inputs for entities) relative to model E (dedicated embedding for each entity);
(b) add cross-lingual entity-entity task on top of the mention-entity task for model F;
(c) control label balance per-entity during negative mining (versus not).
\label{tab:three_pairwise}}
\end{table*}

\subsubsection{Entity Encoder Comparison}
We first consider the choice of entity encoder, comparing Model F with respect to Model E.

\autoref{tab:three_pairwise}(a) shows that using the entity descriptions as inputs leads to dramatically better performance on rare and unseen entities, in exchange for small losses on entities appearing more than 100 times, and overall improvements in both macro and micro recall.

Note that as expected, the embedding Model~E
gives 0\% recall in zero-shot cases, 
as their embeddings are randomly
initialized and never get updated in absence of any training examples.

The embedding table of Model E has 6 billion parameters, but there is no sharing across entities.
Model F has approximately 50 times fewer parameters, but can distribute information in its shared, compact WordPiece vocabulary and Transformer layer parameters.
We can think of these dual encoder models as classifiers over 20 million classes where the softmax layer is either parameterized by an ID embedding (Model E) or an encoding of a description of the class itself (Model F).
Remarkably, using a Transformer for the latter approach effectively compresses (nearly) all the information in the traditional embedding model into a compact and far more generalizable model.

This result highlights the value of analyzing model behavior in terms of entity frequency.
When looking at the micro-averaged metric in isolation, one might conclude that the two models perform similarly;
but the macro-average is sensitive to the large differences in the low-frequency bins.

\subsubsection{Auxiliary Cross-Lingual Task}
In seeking to improve the performance of Model~F on tail entities, we return to the (partly redundant) entity descriptions in multiple languages.
By choosing just one language as the input, we are ignoring potentially valuable information in the remaining descriptions.

Here we add an auxiliary task: cross-lingual entity description retrieval.
This reuses the entity encoder $\psi$ of Model F to map two descriptions of an entity $e$ to a score, $s(t_e^l,t_e^{l'})\propto \psi(t_e^l)^T\psi(t_e^{l'})$,
where $t_e^{l'}$ is the description selected by the earlier heuristic,
and $t_e^l$ is sampled from the other available descriptions for the entity.

We sample up to 5 such cross-lingual pairs per entity to construct the training set for this auxiliary task.
This makes richer use of the available multilingual descriptions, and exposes the model to
39 million additional high-quality training examples whose distribution is decoupled from that of the mention-entity pairs in the primary task.
The multi-task training computes an overall loss by averaging the in-batch sampled softmax loss for a batch of $(m,e)$ pairs and for a batch of $(e,e)$ pairs.

\autoref{tab:three_pairwise}(b) confirms this brings consistent quality gains across all frequency bins, and more so for uncommon entities. Again, reliance on the micro-average metric alone understates the benefit in this data augmentation step for rarer entities.

\begin{figure*}
  \centering
  \includegraphics[width=\textwidth]{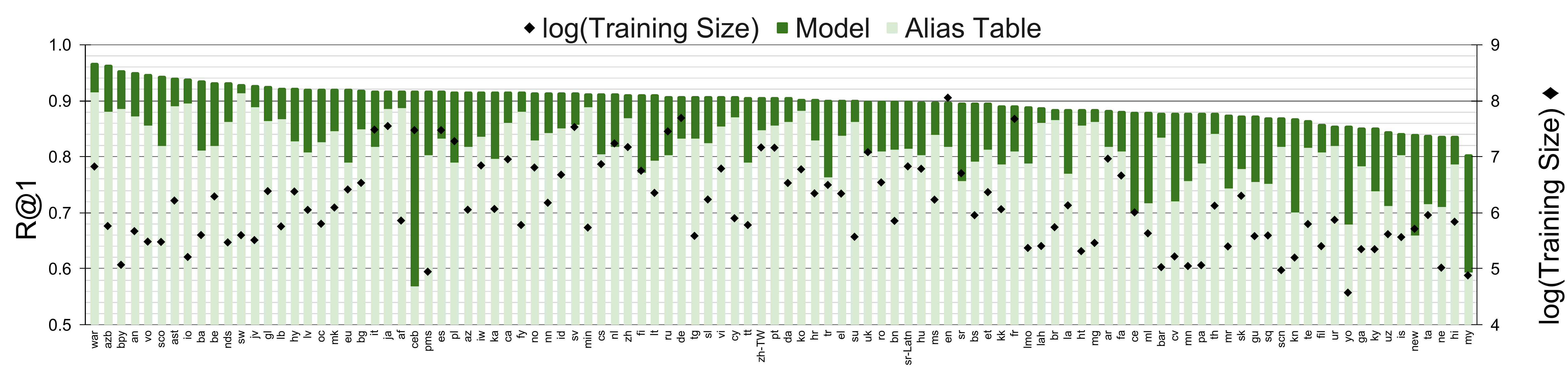}
\vspace{-2.8em}
  \caption{Accuracy of Model F\textsuperscript{+} on the 104 languages in our balanced Wikipedia heldout set, overlayed on alias table accuracy and Wikipedia training set size. (See \autoref{fig:heldout_104_larger} in the Appendix for a larger view.)
  \label{fig:heldout_104}}
\end{figure*}

\subsubsection{Hard-Negative Mining}
Training with hard-negatives is highly effective in monolingual entity retrieval
\citep{gillick-etal-2019-learning}, and we apply the technique they detail to our multilingual setting.

In its standard form, a certain number of negatives are mined for each mention in the training set by collecting top-ranked but incorrect entities retrieved by a prior model.
However, this process can lead to a form of the class imbalance problem as uncommon entities become over-represented as negatives in the resulting data set.
For example, an entity appearing just once in the original training set could appear hundreds or thousands of times as a negative example.
Instead, we control the ratio of positives to negatives on a per-entity basis, mining up to 10 negatives per positive.

\autoref{tab:three_pairwise}(c)
confirms that our strategy effectively addresses the imbalance issue for rare entities with only small degradation for more common entities.
We use this model to perform a second, final round
of the adapted negative mining followed by further training to improve on the macro-average further by +.05 (holdout) and +.08 (\TR{}).

The model we use in the remainder of the experiments combines all these findings. We use Model F with the entity-entity auxiliary task and hard negative mining with per-entity label balancing, referenced as
\textbf{Model F\textsuperscript{+}}.

\subsection{Linking in 100 Languages}\label{sec:linking_in_100}
Breaking down the model's performance by language (R@1 on our heldout set) reveals relatively strong performance across all languages,
despite greatly varying training sizes (\autoref{fig:heldout_104}).
It also shows improvement over an alias table baseline on all languages.
While this does not capture the relative difficulty of the EL task in each language, it does strongly suggest effective cross-lingual transfer in our model: even the most data-poor languages have reasonable results.
This validates our massively multilingual approach.

\subsection{Comparison to Prior Work}\label{sec:eval_compare_to_prior_work}
We evaluate the performance of our final retrieval model relative to previous work on two existing datasets,
noting that direct comparison is impossible because our task setting is novel.

\subsubsection{Cross-Lingual Wikification Setting}\label{sec:eval_xel}
We compare to two previously reported results on \TR:
the \textsc{WikiME} model of \newcite{tsai-roth-2016-cross} that accompanied the dataset, and
the \textsc{Xelms-multi} model by \newcite{upadhyay-etal-2018-joint}.
Both models depend at their core on multilingual word embeddings, which are obtained by applying (bilingual) alignment or projection techniques to pretrained monolingual word embeddings.

\begin{table}
\small
\centering
\begin{tabular}{r cccc} \toprule
 & \textbf{Tsai+} & \textbf{Upad.+} & \textbf{Model F\textsuperscript{+}}\\ \midrule
\textbf{Languages} & 13 & 5 & 104 \\ 
\textbf{$|V|$} & 5m & 5m & 20m \\
\textbf{Candidates}  & 20 & 20 & 20m \\ \midrule
de & 0.53 & 0.55 & \textbf{0.62} \\
es & 0.54 & 0.57 & \textbf{0.58} \\
fr & 0.48 & 0.51 & \textbf{0.54} \\
it & 0.48 & 0.52 & \textbf{0.56} \\ \midrule
\textbf{Average} & 0.51 & 0.54 & \textbf{0.57} \\ \bottomrule
\end{tabular}
\caption{Our best model outperforms previous related non-monolingual models that relied on alias tables and disambiguated among a much smaller set of entities.
\emph{Bottom half:} linking accuracy on the \TR{} test set. 
\emph{Top half:} language coverage; entity vocabulary size; and entities disambiguated among at inference time.
\emph{Middle columns:} \citep{tsai-roth-2016-cross} and \citep{ upadhyay-etal-2018-joint}.
\label{tab:sota}}
\end{table}

As reported in \autoref{tab:sota}, our multilingual dual encoder outperforms the other two by a significant margin.
To the best of our knowledge, this is the highest accuracy to-date on this challenging evaluation set. (Our comparison is limited to the four languages on which \newcite{upadhyay-etal-2018-joint} evaluated their multilingual model.)

This is a strong validation of the proposed approach because the experimental setting is heavily skewed toward the prior models:
Both are \emph{rerankers}, and require a first-stage candidate generation step.
They therefore only disambiguate among the resulting $\leq$20 candidate entities (only from English Wikipedia), whereas our model performs retrieval against all 20 million entities.

\begin{table}
\centering
\begin{tabular}{r cc} \toprule
 & \textbf{DEER} &  \textbf{Model F\textsuperscript{+}} \\ \midrule
\textbf{Languages} & 1 & 104 \\ 
\textbf{Candidates = $|V|$} & 5.7m & 20m \\ \midrule
R@1    & 0.92 & 0.92 \\
R@100  & 0.98 & \textbf{0.99} \\ \bottomrule
\end{tabular}
\caption{Comparison to DEER model \citep{gillick-etal-2019-learning} on their English WikiNews-2018 dataset.
\label{tab:flare_wikinews}}
\end{table}

\subsubsection{Out-of-Domain English Evaluation}\label{sec:eval_english}
We now turn to the question of how well the proposed multilingual model can maintain competitive performance in English and generalize to a domain other than Wikipedia.
\newcite{gillick-etal-2019-learning} provides a suitable comparison point.
Their DEER model is closely related to our approach, but used a more light-weight dual encoder architecture with bags-of-embeddings and feed-forward layers without attention and was evaluated on English EL only.
On the English WikiNews-2018 dataset they introduced, our Transformer-based multilingual dual encoder matches their monolingual model's performance at R@1 and improves R@100 by 0.01 (reaching 0.99)
Our model thus retains strong English performance despite covering many languages and linking against a larger KB. See \autoref{tab:flare_wikinews}.

\subsection{Evaluation on Mewsli-9}\label{sec:eval_wikinews9}
\autoref{tab:wn9_results_main} shows the performance of our model on our new Mewsli-9 dataset compared with an alias table baseline that retrieves entities based on the prior probability of an entity given the observed mention string.
\autoref{tab:wn9_results_bin} shows the usual frequency-binned evaluation. While overall (micro-average) performance is strong, there is plenty of headroom in zero- and few-shot retrieval.

\begin{table}[t]
\centering
\begin{tabular}{r rrrr} \toprule
& \multicolumn{2}{c}{\bf Alias Table} & \multicolumn{2}{c}{\bf Model F\textsuperscript{+}} \\ \cmidrule(lr){2-3} \cmidrule(lr){4-5}
\textbf{Language} & R@1 & R@10 & R@1 & R@10 \\ \midrule
ar & 0.89 & 0.93 & 0.92 & 0.98 \\
de & 0.86 & 0.91 & 0.92 & 0.97 \\
en & 0.79 & 0.86 & 0.87 & 0.94 \\
es & 0.82 & 0.90 & 0.89 & 0.97 \\
fa & 0.87 & 0.92 & 0.92 & 0.97 \\
ja & 0.82 & 0.90 & 0.88 & 0.96 \\
sr & 0.87 & 0.92 & 0.93 & 0.98 \\
ta & 0.79 & 0.85 & 0.88 & 0.97 \\
tr & 0.80 & 0.88 & 0.88 & 0.97 \\ \midrule
micro-avg & 0.83 & 0.89 & 0.89 & 0.96 \\
macro-avg & 0.83 & 0.89 & 0.90 & 0.97 \\ \bottomrule
\end{tabular}
\caption{Results of our main dual encoder Model~F\textsuperscript{+} on the new Mewsli-9 dataset.
Consistent performance across languages in a different domain from the training set points at good generalization.
\label{tab:wn9_results_main}}
\end{table}

\begin{table}[t]
\small
\centering
\begin{tabular}{l rrr r} \toprule
& & \multicolumn{2}{c}{\bf Model F\textsuperscript{+}} & \multicolumn{1}{c}{\bf +CA} \\ 
\cmidrule(lr){3-4}
\cmidrule(lr){5-5}
\textbf{Bin} & \textbf{Queries}  & \emph{R@1} & R@10 & \emph{R@1} \\
\midrule
$[0, 1)$ & 3,198 & 0.08 & 0.34 & 0.07 \\
$[1, 10)$  & 6,564 & 0.58 & 0.81 & 0.60 \\
$[10, 100)$ & 32,371 & 0.80 & 0.93 & 0.82 \\
$[100, 1k)$ & 66,232 & 0.90 & 0.97 & 0.90 \\
$[1k, 10k)$ & 78,519 & 0.93 & 0.98 & 0.93 \\
$[10k, +)$  & 102,203 & 0.94 & 0.99 & 0.96 \\ \midrule
micro-avg & 289,087 & 0.89 & 0.96 & 0.91 \\
macro-avg &  & 0.70 & 0.84 & 0.71 \\ \bottomrule
\end{tabular}
\caption{Results on the new Mewsli-9 dataset, by entity frequency, attained by our main dual encoder Model~F\textsuperscript{+}, plus reranking its predictions with a Cross-Attention scoring model (CA).
\label{tab:wn9_results_bin}}
\end{table}

\begin{table*}
\small
\centering
\begin{tabular}{r L{.86\textwidth}} \toprule
  \textbf{Context 1} & \ldots Bei den neuen Bahnen handelt es sich um das Model \textbf{Tramino} von der polnischen Firma Solaris Bus \& Coach\ldots \\ \cmidrule(lr){2-2}
  \textbf{Prediction} & \ientity{Solaris Tramino}{Q780281}: Solaris Tramino -- rodzina tramwajów, które są produkowane przez firmę Solaris Bus \& Coach z Bolechowa koło Poznania\ldots \\ \cmidrule(lr){2-2}
  \textbf{Outcome} & \emph{\textbf{Correct}: A family of trams originally manufactured in Poland, mentioned here in German, linked to its Polish description.} \\ \midrule
      
  \textbf{Context 2} & \ldots sobre una tecnología que permitiría fabricar chocolate a partir de los \textbf{zumos de fruta}, agua con vitamina C o gaseosa dietética\ldots   \\ \cmidrule(lr){2-2}
  \textbf{Prediction} & \ientity{fruit juice}{Q20932605}: Fruchtsaft , spezieller auch Obstsaft , ist ein aus Früchten einer oder mehrerer Fruchtarten gewonnenes flüssiges Erzeugnis\ldots  \\ \cmidrule(lr){2-2}
  \textbf{Outcome} & \emph{\textbf{Correct}: A Spanish mention of ``fruit juice'' linked to its German description---only ``juice'' has a dedicated English Wikipedia page.} \\ \midrule

  \textbf{Context 3} & 
  \foreignlanguage{russian}{\ldots Душан Ивковић рекао је да је његов тим имао императив победе над ( Италијом ) на Европском првенству\ldots } \\  \cmidrule(lr){2-2}
  \textbf{Prediction} & \ientity{It.\ men's water polo team}{Q261190}: La nazionale di pallanuoto maschile dell' Italia\ldots  \\ \cmidrule(lr){2-2}
  \textbf{Expected} & \ientity{It.\ nat.\ basketball team}{Q261190}: La nazionale di pallacanestro italiana è la selezione dei migliori giocatori di nazionalità italiana\ldots  \\ \cmidrule(lr){2-2}
  \textbf{Outcome} & \emph{\textbf{Wrong}: A legitimately ambiguous mention of ``Italy'' in Serbian (sports context),
  for which model retrieved the water polo and football teams, followed by the expected basketball team entity, all featurized in Italian.} 
  \\\midrule

  \textbf{Context 4} & 
      \ldots In July 2009 , action by the Federal Bureau of Reclamation to protect threatened fish stopped \textbf{irrigation pumping} to parts of the California Central Valley\ldots  \\ \cmidrule(lr){2-2}
  \textbf{Prediction} &
      \ientity{irrigation sprinkler}{Q998539}: \begin{CJK}{UTF8}{min}スプリンクラー は 、 水 に 高圧 を かけ 飛 沫 に し て ノズル から 散布 する 装置\end{CJK} \\ \cmidrule(lr){2-2}
  \textbf{Outcome} & \emph{\textbf{Wrong}: Metonymous mention of \ientity{Central Valley Project}{Q2944429} in English, but model retrieved the more literal match, featurized in Chinese. Metonymy is a known challenging case for EL \citep{ling2015design}.} \\ \bottomrule
%  %Other languages work like this: (configure 'babel' at the top of the file)
%  %\foreignlanguage{russian}{Джамахири́я}
\end{tabular}
\caption{Correct and mistaken examples observed in error analysis of dual encoder model F\textsuperscript{+} on Mewsli-9.\label{fig:examples}}
\end{table*}

\subsubsection{Example Outputs}
We sampled the model's correct predictions on Mewsli-9, focusing on cross-lingual examples where entities do not have an English Wikipedia page (\autoref{fig:examples}). These examples demonstrate that the model effectively learns cross-lingual entity representations.
Based on a random sample of the model's errors, we also show examples that summarize notable error categories.

\subsubsection{Reranking Experiment}
We finally report a preliminary experiment to apply a cross-attention scoring model (CA) to rerank entity candidates retrieved by the main dual encoder (DE), using the same architecture of \newcite{logeswaran-etal-2019-zero}. We feed the concatenated mention text and entity description into a 12-layer Transformer model, initialized from the same multilingual BERT checkpoint referenced earlier.

The CA model's CLS token encoding is used to classify mention-entity coherence. We train the model with a binary cross-entropy loss, using positives from our Wikipedia training data, taking for each one the top-4 DE-retrieved candidates plus 4 random candidates 
(proportional to the positive distributions).

We use the trained CA model to rerank the \mbox{top-5} DE candidates for Mewsli-9 (\autoref{tab:wn9_results_bin}). We observed improvements on most frequency buckets compared to DE R@1, which suggests that the model's few-shot capability can be improved by cross-lingual reading-comprehension.
This also offers an initial multilingual validation of a similar two-step BERT-based approach recently introduced in a monolingual setting by
\citep{wu2019zeroshot}, and provides a strong baseline for future work.

\section{Conclusion}
We have proposed a new formulation for multilingual entity linking that seeks to expand the scope of entity linking to better reflect the real-world challenges
of rare entities and/or low resource languages.
Operationalized through Wikipedia and WikiData, our experiments using enhanced dual encoder retrieval models and frequency-based evaluation provide compelling evidence that it is feasible to perform this task with a single model covering over a 100 languages.

Our automatically extracted Mewsli-9 dataset serves as a starting point for evaluating entity linking beyond the entrenched English benchmarks and under the expanded multilingual setting.
Future work could investigate the use of non-expert human raters to improve the dataset quality further.

In pursuit of improved entity representations, future work could explore the joint use of complementary multi-language descriptions per entity, methods to update representations in a light-weight fashion when descriptions change, and incorporate relational information stored in the KB.

\section*{Acknowledgments}
Thanks to Sayali Kulkarni and the anonymous reviewers for their helpful feedback.

\bibliographystyle{acl_natbib}
\bibliography{emnlp2020}

\appendix

\clearpage

\renewcommand{\thesubsection}{\Alph{subsection}}
\section*{Appendix}
\subsection{Mewsli-9 Dataset}\label{appsec:mewsli}
\emph{Available at:} \url{http://goo.gle/mewsli-dataset}

We used an automated process to construct Mewsli-9, exploiting link anchor text to identify naturally occurring entity mentions in WikiNews articles, from its inception to the end of 2018.

From a given WikiNews page dump,\footnote{%
\url{archive.org/download/XXwikinews-20190101/XXwikinews-20190101-pages-articles.xml.bz2}
where \texttt{XX} is a language code.
}
we extracted text including link anchors and section headings using a modified version of \texttt{\small wikiextractor}.\footnote{\url{github.com/attardi/wikiextractor}}

To obtain clean article text, we discard page-final sections that merely contain external references, etc.
This is done by matching section headings against a small set of hand-collected, language-specific patterns.

Mention candidates are filtered to those remaining links that point to Wikipedia pages in any language (not limited to our 104 languages).
These Wikipedia links are redirected if necessary, and resolved to WikiData identifiers to determine the gold entity for a mention.
There are many reasons why resolution may fail, including mistakes in the original markup and churn in the data sources over time.
The final dataset is limited to (mention, entity) pairs where resolution to WikiData succeeded.

\subsection{Training Details and Hyperparameters}\label{appsec:training}
All model training was carried out on a Google TPU v3 architecture,\footnote{\url{cloud.google.com/tpu/docs/tpus}} using batch size 8192 and
a learning rate schedule that uses linear warm-up followed by linear decay to 0.

The first phase of training our dual encoders (DE) with in-batch random negatives encompasses 500,000 steps, which takes approximately one day.

Where hard-negative training is applied, we initialize from the corresponding prior model checkpoint
and continue training against the multi-task loss for a further 250,000 steps, which also takes about a day.

Other than the limit to using a 4-layer Transformer stack, our mention encoder and model F entity encoders use the same hyperparameters as mBERT-base, allowing initialization from the publicly available checkpoint---we use the weights of its first 4 layers, in addition to those of the token and positional embeddings.

The cross-attention scoring model (CA) in the final preliminary experiment is a full 12-layer Transformer (also mBERT-base), and was trained for 1 million steps, taking just under one day.

The learning rates were 1e-4 (DE) and 1-e5 (CA) and included warm-up phases of 10\% (DE) and 1\% (CA) of the respective number of training steps.

\subsection{Data Preprocessing}\label{appsec:dataprep}
We used the 2019-10-03 dump of Wikipedia and WikiData, parsed using in-house tools.\footnote{\url{dumps.wikimedia.org}}

Two filtering criteria are relevant in preprocessing WikiData to define our KB. 
The first is to exclude items that are a subclass (P279) or instance of (P31) the most common Wikimedia-internal administrative entities, detailed in \autoref{tab:wikidata_filter}.
The remaining entities are then filtered to retain only those for which the WikiData entry points to at least one Wikipedia page, in any language, motivated by our objective of using descriptive text as entity features.

\begin{table}
\centering
\begin{tabular}{l l} \toprule
{\bf QID} & {\bf label} \\ \midrule
Q4167836  &  category \\
Q24046192 &  category stub \\
Q20010800 &  user category \\
Q11266439 &  template \\
Q11753321 &  navigational template \\
Q19842659 &  user template \\
Q21528878 &  redirect page \\
Q17362920 &  duplicated page \\
Q14204246 &  project page \\
Q21025364 &  project page \\
Q17442446 &  internal item \\
Q26267864 &  KML file \\
Q4663903  &  portal \\
Q15184295 &  module \\ \bottomrule
\end{tabular}
\caption{WikiData identifiers used for filtering out Wikimedia-internal entities from our KB definition.
\label{tab:wikidata_filter}}
\end{table}

\subsection{Other data sources}\label{appsep:otherdata}
-- \TR{} dataset:
\url{cogcomp.seas.upenn.edu/page/resource_view/102}
%\url{bilbo.cs.illinois.edu/~ctsai12/xlwikifier-wikidata.zip}.
%now seems dead but pretty sure that's where I got it in 2019H1

\noindent -- English WikiNews-2018 dataset by \cite{gillick-etal-2019-learning}: \url{github.com/google-research/google-research/tree/master/dense_representations_for_entity_retrieval}
        
\renewcommand{\thefigure}{B\arabic{figure}}
\setcounter{figure}{0}
\begin{sidewaysfigure*}
  \centering
  \includegraphics[width=\textwidth]{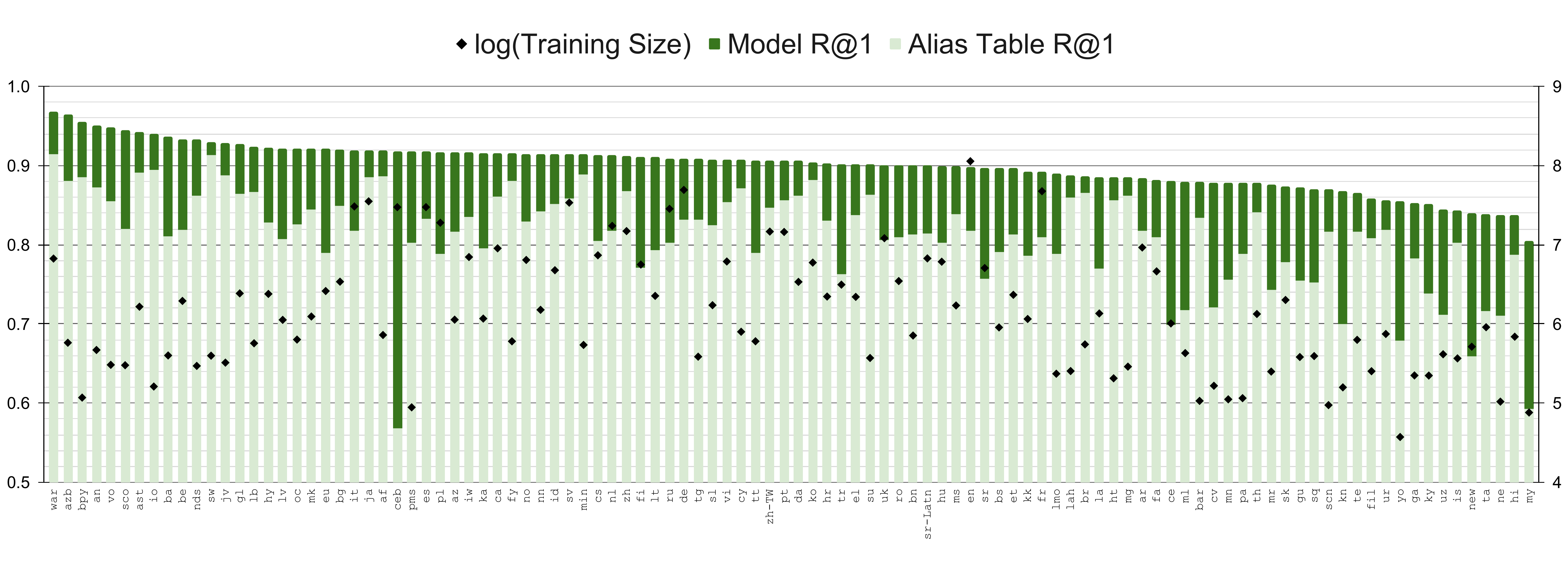}
  \caption{ Accuracy (left-axis) of Model F\textsuperscript{+} on the 104 languages in our balanced Wikipedia heldout set, overlayed on alias table accuracy, and Wikipedia training set size (right-axis).
  Our Model F\textsuperscript{+} obtains relatively strong performance across 104 languages and outperforms an alias table baseline in all cases.
  Although task difficulty is not necessarily comparable between languages, this result suggests effective cross-lingual transfer is happening: even languages with small training sets show reasonable accuracy.
  \emph{(This is a scaled-up reproduction of \autoref{fig:heldout_104} from \autoref{sec:linking_in_100}.)}
  \label{fig:heldout_104_larger}}
\end{sidewaysfigure*}
        
%%%%%%%%%%%%%% TABLE TEMPLATE
%\begin{table}
%\centering
%\begin{tabular}{c|c} \toprule
%&  \\ \midrule
%&  \\ \bottomrule
%\end{tabular}
%\caption{Caption.
%\label{tab:REFREF}}
%\end{table}
%%%%%%%%%%%%%% TABLE TEMPLATE

\end{document}